# Measuring Inconsistency in Probabilistic Knowledge Bases

Matthias Thimm
Information Engineering Group, Department of Computer Science,
Technische Universität Dortmund, Germany

## Abstract

This paper develops an inconsistency measure on conditional probabilistic knowledge bases. The measure is based on fundamental principles for inconsistency measures and thus provides a solid theoretical framework for the treatment of inconsistencies in probabilistic expert systems. We illustrate its usefulness and immediate application on several examples and present some formal results. Building on this measure we use the Shapley value—a well-known solution for coalition games—to define a sophisticated indicator that is not only able to measure inconsistencies but to reveal the causes of inconsistencies in the knowledge base. Altogether these tools guide the knowledge engineer in his aim to restore consistency and therefore enable him to build a consistent and usable knowledge base that can be employed in probabilistic expert systems.

## 1   Introduction

Inconsistencies arise easily when experts share their knowledge in order to build a common knowledge base. Although these inconsistencies often affect only a little portion of the knowledge base or emerge from only little differences in the experts knowledge, they cause severe damage. Especially in knowledge bases that use classical logic as a means for knowledge representation, inconsistencies render the whole knowledge base useless, due to the well-known principle *ex falso quodlibet*. Therefore reasoning under inconsistency is an important field in AI and there are many proposals to deal with inconsistency in classical logic, e.g. (Rescher and Manor, 1970; Konieczny *et al.*, 2005), or in other logical frameworks, e.g. paraconsistent logics (Bziau *et al.*, 2007), default logics (Reiter, 1980), defeasible logics (Billington, 2008), and argumentation theory (Bench-Capon and Dunne, 2007). Furthermore there are several approaches to analyze and measure inconsistency in qualitative frameworks, e.g. (Lozinskii, 1994; Benferhat *et al.*, 1997; Knight, 2001; Hunter and Konieczny, 2004), and some in quantitative frameworks (mainly possibilistic frameworks), e.g. (Dubois *et al.*, 1992).

Here, we aim at analyzing inconsistencies in a probabilistic framework and in particular measuring inconsistency in conditional probabilistic knowledge bases (Nute and Cross, 2002; Kern-Isberner, 2001; Benferhat *et al.*, 1999; Rödder and Meyer, 1996). In these, knowledge is captured using conditionals $(A \mid B)$ that describe rules of the form *"If B then A"* and are interpreted using conditional probabilities. In contrast to probabilistic networks like Bayesian Networks (Pearl, 1998) conditional probabilistic knowledge bases do not demand the complete specification of every conditional probability of every probabilistic dependence and thus do not define a unique probability distribution as the underlying model. Nonetheless, using maximum entropy methods (Grove *et al.*, 1994) one can determine a single probability distribution that describes the specified knowledge in an unbiased way. However, due to the unstructured approach of conditional probabilistic knowledge bases inconsistencies easily occur whereas Bayesian Networks forbid cyclic dependencies and so inconsistencies cannot arise by definition.

There is very little work on the treatment of inconsistencies in a conditional probabilistic framework (Rödder and Xu, 2001; Finthammer *et al.*, 2007). The method described in (Finthammer *et al.*, 2007) consists of a set of heuristics that are used to restore consistency in a knowledge base. Although this method is not based on a theoretical elaboration it works well in real-world examples and has been applied successfully to improve fraud detection in management. Other related work (Hansen and Jaumard, 1996; Andersen and Pretolani, 2001) investigates inconsistencies in classi-cal theories enriched with probabilistic semantics but without treatment of conditional probabilities as we do here. Furthermore the authors of (Hansen and Jaumard, 1996; Andersen and Pretolani, 2001) are mainly interested in efficient algorithms for determining *whether* a knowledge base is inconsistent and not to what degree.

The main contribution of this paper consists of the development of an inconsistency measure on conditional probabilistic knowledge bases that is based on fundamental principles that are desired from inconsistency measures in general. Furthermore we apply the approach taken in (Hunter and Konieczny, 2006) on our framework and use the Shapley value (Shapley, 1953) to define a more sophisticated inconsistency measure based on the basic inconsistency measure developed before. Using this Shapley inconsistency measure we are able to determine not only the degree of inconsistency but also the contributors to it, i.e. the conditionals that are responsible for the inconsistency.

This paper is organized as follows. In Section 2 we start with some necessary preliminaries and a description of the basic framework. Afterwards in Section 3 we discuss the basic problem of determining consistency in the basic framework. In Section 4 we go on by stating some desirable properties of an inconsistency measure on probabilistic knowledge bases in general and propose a measure that fulfills these properties. In order to be able to determine the causes of inconsistency we continue in Section 5 by introducing a more sophisticated measure based on the Shapley value and the basic inconsistency measure introduced before. In Section 6 we conclude.

## 2　Preliminaries

We are working with a propositional framework of random variables. Let $\mathbf{V} = \{V_1, \ldots, V_n\}$ be a set of propositional variables with finite domains $\mathsf{Dom}(V_1), \ldots, \mathsf{Dom}(V_n)$. The set $\mathbf{V}$ is assumed to be given for all upcoming definitions. An expression of the form $V_i = v_i$ is called a *literal* if $v_i$ is in the domain of $V_i$, i.e. $v_i \in \mathsf{Dom}(V_i)$. The language $\mathcal{L}_\mathbf{V}$ is generated using the connectives $\neg, \wedge$, and $\vee$ on the literals in $\mathbf{V}$ in the usual way. We abbreviate conjunctions $A \wedge B$ by $AB$ and negation $\neg A$ by overlining $\overline{A}$. If $V$ is a binary variable, i.e., it is $\mathsf{Dom}(V) = \{\mathsf{true}, \mathsf{false}\}$, we abbreviate $V = \mathsf{true}$ by just $V$ and $V = \mathsf{false}$ by $\overline{V}$. We write $\top$ for tautological formulas, e.g. $A \vee \overline{A} = \top$. A *complete conjunction* or interpretation is a conjunction of literals where every $V_i \in \mathbf{V}$ appears exactly once. If $\omega$ is a complete conjunction, then $\omega \models (V_i = v_i)$ if and only of $V_i = v_i$ appears in $\omega$. For an arbitrary formula $B$ the expression $\omega \models B$ evaluates in the usual way.

Let $\Omega$ be the set of all complete conjunctions of $\mathbf{V}$, i.e., the set of all interpretations of the propositional language induced by $\mathbf{V}$.

Probabilistic knowledge bases are build using probabilistic constraints, that impose certain restrictions on the conditional probabilities of the models of the knowledge base. A *probabilistic constraint* $r$ is an expression of the form $(A \,|\, B)[d]$ with formulas $A, B$ and $d \in [0, 1]$. If $B = \top$ we write $(A)[d]$ instead of $(A \,|\, \top)[d]$. A set of probabilistic constraints $R = \{r_1, \ldots, r_m\}$ is called a *knowledge base*. Let $\mathfrak{R}$ denote the set of all knowledge bases. The models of a knowledge base $R$ are the probability distributions $P_R : \Omega \to [0, 1]$ that fulfill all restrictions on the conditional probabilities imposed by the probabilistic constraints in $R$. More specific, a probability distribution $P_R : \Omega \to [0, 1]$ is a model for a knowledge base $R$, written $P_R \models R$, if and only if $P_R \models r$ for every $r \in R$. That is

$$P_R \models (A \,|\, B)[d] \quad :\Leftrightarrow \quad P_R(A \,|\, B) = d$$
$$\Leftrightarrow \quad P_R(AB) = d \cdot P(B)$$

with

$$P_R(A) = \sum_{\omega \in \Omega, \omega \models A} P_R(\omega) \quad .$$

Observe, that we used the notation $P_R(AB) = d \cdot P(B)$ to express that the conditional probability of $P_R(A \,|\, B)$ is $d$ in order to avoid a case differentiation for $P_R(B) = 0$. For the rest of this paper, we assume that all probabilistic constraints are *self-consistent*, i.e., for every singleton set $\{r\}$ with $r$ being a probabilistic constraint we assume that $\{r\}$ has a model. For example, we forbid constraints of the form $(A \,|\, \overline{A})[d]$ with $d > 0$ and the like.

A knowledge base $R$ made of probabilistic constraints describes incomplete knowledge. Usually, one is interested in performing inductive representation techniques and thus in computing a single probability distribution that describes $R$ best and thus gives a complete description of the problem area at hand. This can be done using methods based on maximum entropy, which feature several nice properties (Paris, 1994; Grove *et al.*, 1994; Kern-Isberner, 2001; Rödder and Meyer, 1996). Although these methods are not the topic of the present paper, consistency is a necessary requirement for their application.

## 3　Determining Consistency

It must not always be the case, that a probability distribution $P_R$ with $P_R \models R$ for a knowledge base $R$ exists, e.g., for the knowledge base $R = \{(A \,|\, B)[0.5], (B \,|\, \top)[0.5], (A \,|\, \top)[0.1]\}$ with literals $A, B$ there is no distribution $P_R$ with $P_R \models R$.

Hence, we are interested in determining for a specific knowledge base $R$, whether $R$ is consistent (if there is at least on distribution $P_R$ with $P_R \models R$) or inconsistent (if there is no such $P_R$).

In the following, we reduce this problem to a constraint satisfaction problem similar to the approaches in (Hansen and Jaumard, 1996; Andersen and Pretolani, 2001), which consider purely propositional knowledge bases without conditional constraints. Let $R$ be a knowledge base. For a probability distribution $P_R : \Omega \to [0,1]$ to be a model of $R$, every probabilistic constraint $(A \mid B)[d] \in R$ imposes $P_R(A \mid B) = d$ to hold. Let $Mod(A) = \{\omega \in \Omega \mid \omega \models A\}$ be the set of all models of the formula $A$. For every complete conjunction $\omega \in \Omega$ we introduce a variable $\alpha_\omega$ that determines the unknown value of $P_R(\omega)$. Then $(A \mid B)[d]$ translates to

$$\sum_{\omega \in Mod(AB)} \alpha_\omega = d \cdot \sum_{\omega \in Mod(B)} \alpha_\omega . \qquad (1)$$

In order to ensure that $P_R$ is indeed a probability distribution we need the following normalization constraints

$$\sum_{\omega \in \Omega} \alpha_\omega = 1 \qquad (2)$$

$$\forall \omega \in \Omega : \alpha_\omega \geq 0 \quad . \qquad (3)$$

Taken together for all probabilistic constraints $r \in R$ the corresponding equation (1) and the equations (2) and (3), this yields a constraint satisfaction problem $CS_R$ on the variables $\{\alpha_\omega \mid \omega \in \Omega\}$.

**Proposition 1.** *Let $R$ be a knowledge base. $R$ is consistent if and only if $CS_R$ has a solution.*

The proof of Proposition 1 is straightforward as every assignment of values to the variables $\alpha_\omega$, that is legal with respect to the constraint satisfaction problem $CS_R$, directly corresponds to a probability distribution $P_R$ with $P_R(\omega) = \alpha_\omega$. Hence, if there is an assignment for all $\alpha_\omega$ the corresponding probability distribution $P_R$ is a model for all probabilistic constraints $r \in R$ and therefore a model for $R$.

## 4 Measuring Inconsistency

The simple piece of information that a knowledge base $R$ is inconsistent is not always sufficient for knowledge engineering and analyzing. In order to fix the knowledge base more detailed information on the inconsistency is necessary. There is much work on analyzing inconsistency in qualitative frameworks, see e.g. (Knight, 2001; Wong and Besnard, 2001; Hunter and Konieczny, 2006), but there is very less work on analyzing inconsistency in quantitative frameworks, especially in probabilistic frameworks as the one discussed here (Finthammer *et al.*, 2007). While (Finthammer *et al.*, 2007) is mainly concerned with resolving inconsistencies using heuristics, here we take a more formal approach in the analysis of inconsistency by formalizing and developing an inconsistency measure on probabilistic knowledge bases.

We go on by stating some desirable properties of an inconsistency measure on knowledge bases. Afterwards we propose a simple inconsistency measure that is based on the constraint satisfaction problem $CS_R$ and also fulfills the desirable properties.

### 4.1 Desirable Properties

Let Inc be a function $\mathsf{Inc} : \mathfrak{R} \to [0, \infty]$ that maps a knowledge base $R \in \mathfrak{R}$ onto a positive real. We desire several properties of Inc in order of Inc describing an inconsistency measure. Some of the following properties are adapted from (Hunter and Konieczny, 2006) and rewritten to fit a probabilistic framework. Intuitively we want Inc to be a function on knowledge bases that is monotonically increasing with the inconsistency in the knowledge base. If the knowledge base is consistent, Inc shall be minimal. For the upcoming definitions let $R$ be an arbitrary knowledge base and $r, r'$ be probabilistic constraints.

**(Consistency)** If $R$ is consistent, then $\mathsf{Inc}(R) = 0$.

**(Inconsistency)** If $R$ is inconsistent, then $\mathsf{Inc}(R) > 0$.

The above properties ensure that Inc is indeed an inconsistency measure and not an information measure (Cover, 2001) as it should not distinguish between different consistent knowledge bases but measure inconsistent ones.

**(Monotonicity)** It is $\mathsf{Inc}(R) \leq \mathsf{Inc}(R \cup \{r\})$.

**(Super-Additivity)** If $R \cap R' = \emptyset$, it is $\mathsf{Inc}(R \cup R') \geq \mathsf{Inc}(R) + \mathsf{Inc}(R')$.

The measure of inconsistency can only increase when new pieces of information are added to the knowledge base. Thus inconsistencies cannot be resolved with new information. (Super-Additivity) is the stronger property, as it can be easily seen that (Super-Additivity) implies (Monotonicity).

**(Weak Independence)** If no literal in $r$ is mentioned in $R$, then it is $\mathsf{Inc}(R) = \mathsf{Inc}(R \cup \{r\})$.

As we assume that all probabilistic constraints are self-consistent, the addition of a constraint not involving any parts of the language mentioned yet shall not lead to an increase in the inconsistency.

We say that $r$ is a *free constraint* iff for every set $R' \subseteq R$ such that $R'$ is inconsistent and minimal with this property, it is $r \notin R'$. Then we can strengthen the above property as follows.

**(Independence)** If $r$ is a free constraint in $R \cup \{r\}$, then it is $\mathsf{Inc}(R) = \mathsf{Inc}(R \cup \{r\})$.

This property ensures that not only constraints that do not use literals previously mentioned cannot increase inconsistency, but also constraints that do not take part in any inconsistency of the knowledge base do so. It is easy to see that satisfaction of the second property implies satisfaction of the first property.

**Proposition 2.** *If* $\mathsf{Inc}$ *satisfies (Independence), then* $\mathsf{Inc}$ *satisfies (Weak Independence).*

The previous two properties describe cases where the inconsistency of a knowledge base should remain constant despite the addition of new information. Conversely, the next property describes a case when the inconsistency should increase.

**(Penalty)** If $r$ is not a free constraint in $R \cup \{r\}$, then it is $\mathsf{Inc}(R) < \mathsf{Inc}(R \cup \{r\})$.

Similar to the motivation for (Independence) we state that if a probabilistic constraint $r$ contributes to a minimal inconsistent subset of the knowledge base, then the inconsistency must be strictly greater than in the knowledge base without $r$.

So far we have not taken into account that we are working in a probabilistic framework. It is hard to grasp in what way the probabilities of the conditionals influence the inconsistency of the whole knowledge base. Consider a knowledge base $R$ and a probabilistic constraint $(A \mid B)[d] \in R$. How should the inconsistency measure $\mathsf{Inc}$ behave when increasing (or decreasing) the value $d$? There is no definite answer to this question as, on the one hand, the inconsistency may vanish because the constraint may become consistent with the rest of the knowledge base, or, on the other hand, the inconsistency may rise because the constraint may remove itself from a "consistent state". But one demand can be made: The change in the measure of inconsistency should be continuous in $d$. If one does only slightly change a given knowledge base, the resulting inconsistency measure should have only changed slightly as well. We formalize this intuition as follows.

**Definition 1** (Characteristic function). Let $R = \{(A_1 \mid B_1)[d_1], \ldots, (A_n \mid B_n)[d_n]\}$ be a knowledge base. The function $\Lambda_R : [0,1]^n \to \mathfrak{R}$ with

$$\Lambda_R(x_1, \ldots, x_n) = \{(A_1 \mid B_1)[x_1], \ldots, (A_n \mid B_n)[x_n]\}$$

is called the *characteristic function* of $R$.

**Definition 2** (Characteristic Inconsistency function). Let $R$ be a knowledge base with $|R| = n$. The function $\theta_{\mathsf{Inc},R} : [0,1]^n \to [0, \infty]$ with $\theta_{\mathsf{Inc},R} = \mathsf{Inc} \circ \Lambda_R$ is called the *characteristic inconsistency function* of $\mathsf{Inc}$ and $R$.

The above definitions allow us to state our last property in a concise way as follows.

**(Continuity)** The characteristic inconsistency function $\theta_{\mathsf{Inc},R}$ is continuous in all arguments.

### 4.2 An Inconsistency Measure

In this subsection we develop an inconsistency measure on probabilistic knowledge bases that fulfills the basic properties described above. For this reason, we extend the consistency test from the previous section by including variables that measure the deviation of the values of the probabilistic constraints from consistent ones in a minimal way. But first, consider the following remark.

**Remark 1.** *For every set* $R' = \{(A_1 \mid B_1), \ldots, (A_n \mid B_n)\}$ *of qualitative conditionals (neglecting the probabilistic values) there are reals* $d_1, \ldots, d_n \in [0,1]$, *such that* $R = \{(A_1 \mid B_1)[d_1], \ldots, (A_n \mid B_n)[d_n]\}$ *is consistent.*

This is easy to see, because one can consider any probability distribution $P$ and assign $d_i := P(A_i \mid B_i)$. As we only consider self-consistent constraints, every value is computable. Bearing this observation in mind, let $R = \{(A_1 \mid B_1)[d_1], \ldots, (A_n \mid B_n)[d_n]\}$ be a knowledge base. For every $i = 1, \ldots, n$ we introduce variables $\eta_i, \tau_i \in [0,1]$ that measure the positive and negative minimal deviations of the value of the probabilistic constraint $(A_i \mid B_i)[d_i]$. In the following, we define an optimization problem, that minimizes the deviation of $R$ to a consistent knowledge base. To this end, we have to modify the probabilistic constraints in $R$ in a minimal way, such that the knowledge base $R'$ with the modified constraints is consistent, i.e., there is a probability distribution $P_{R'}$ that is a model for $R'$. As before let $\alpha_\omega$ denote the probability of a complete conjunction $\omega \in \Omega$. For every constraint $(A_i \mid B_i)[d_i]$, $i = 1, \ldots, n$ we write

$$\sum_{\omega \in Mod(A_i B_i)} \alpha_\omega = (d_i + \eta_i - \tau_i) \cdot \sum_{\omega \in Mod(B_i)} \alpha_\omega \quad (4)$$

to comprehend for the fact that the modified probabilistic constraint $(A_i \,|\, B_i)[d_i + \eta_i - \tau_i]$ imposes $P_{R'}(A_i \,|\, B_i) = d_i + \eta_i - \tau_i$[1]. To ensure well-formed constraints we also have to consider the following normalization constraints

$$0 \leq d_1 + \eta_1 - \tau_1 \leq 1, \ldots, 0 \leq d_n + \eta_n - \tau_n \leq 1 \quad (5)$$

and as before

$$\sum_{\omega \in \Omega} \alpha_\omega = 1 \quad (6)$$

$$\forall \omega \in \Omega : \alpha_\omega \geq 0 \quad . \quad (7)$$

We denote with $OPT_R$ the set of constraints (4), (5), (6), and (7) for a knowledge base $R$. In order to determine the minimal necessary deviation of $R$ from a consistent knowledge base, we formulate an optimization problem by minimizing the function

$$f(\eta_1, \ldots, \eta_n, \tau_1, \ldots, \tau_n) = \eta_1 + \ldots + \eta_n + \tau_1 + \ldots + \tau_n$$

given $OPT_R$. Let $\mathsf{Inc}^*(R)$ denote the solution for $f$ in this optimization problem and let $\eta_1^*, \ldots, \eta_n^*, \tau_1^*, \ldots, \tau_n^*$ be the parameters for a minimal value. Considering again Remark 1, there is always a solution for the optimization problem defined above.

**Proposition 3.** *For every knowledge base $R$, the value $\mathsf{Inc}^*(R)$ is well-defined.*

As not all constraints in this optimization problem are strictly convex, the values of $\eta_1^*, \ldots, \eta_n^*, \tau_1^*, \ldots, \tau_n^*$ do not have to be unique in general (Boyd and Vandenberghe, 2004). But some straightforward observations can be made on the value of $\mathsf{Inc}^*$.

**Proposition 4.** *If $\eta_i^* > 0$ then $\tau_i^* = 0$ and if $\tau_i^* > 0$ then $\eta_i^* = 0$.*

**Proposition 5.** *Let $R = \{(A_1 \,|\, B_1)[d_1], \ldots, (A_n \,|\, B_n)[d_n]\}$ be a knowledge base, then it is*

$$0 \leq \mathsf{Inc}^*(R) \leq \sum_{1 \leq i \leq n} \max(d_i, 1 - d_i) \leq n \quad .$$

For any specific knowledge base $R$ Proposition 5 states that the value of $\mathsf{Inc}^*(R)$ is bounded above by the number of conditionals in $R$. By exploiting this observation one can define a normalized inconsistency measure by

$$\mathsf{Inc}_0^*(R) =_{def} \begin{cases} 0 & \text{if } R = \emptyset \\ \frac{\mathsf{Inc}^*(R)}{|R|} & \text{otherwise} \end{cases}$$

---

[1]We explicitly distinguish between positive ($\eta_i$) and negative ($\tau_i$) deviations to avoid determining absolute values when summing the deviations (see below).

with values between zero and one for any knowledge base $R$. However, we will go on by investigating the properties of the unnormalized inconsistency measure $\mathsf{Inc}^*$, which naturally apply to the normalized inconsistency measure $\mathsf{Inc}_0^*$ as well.

**Example 1.** Consider the knowledge base $R_1 = \{r_1, r_2, r_3, r_4\}$ with $r_1 = (A \,|\, \overline{B})[0.8]$, $r_2 = (A \,|\, B)[0.6]$, $r_3 = (B)[0.5]$, and $r_4 = (A)[0.2]$. Here, it is $\mathsf{Inc}^*(R_1) = 0.5$ with $\eta_1^* = \tau_1^* = \eta_2^* = \tau_2^* = \eta_3^* = \tau_3^* = \tau_4^* = 0$ and $\eta_4^* = 0.5$. Therefore, the fourth constraint $(A)[0.2]$ has to be adjusted to $(A)[0.7]$ in order to restore consistency (this is just one possible adjustment).

**Example 2.** Consider the knowledge base $R_2 = \{r_1, r_2, r_3\}$ with $r_1 = (A \,|\, B)[1]$, $r_2 = (B)[1]$, and $r_3 = (A)[0]$. Here, it is $\mathsf{Inc}^*(R_2) = 1$ with $\eta_1^* = \eta_2^* = \tau_2^* = \eta_3^* = \tau_3^* = 0$ and $\tau_1^* = 1$.

As the next example shows, it must not always be a single probabilistic constraint, that has to be modified in order to restore consistency.

**Example 3.** Consider the knowledge base $R_3 = \{r_1, r_2, r_3, r_4, r_5\}$ with $r_1 = (A \,|\, C)[0.7]$, $r_2 = (B \,|\, \overline{C})[0.8]$, $r_3 = (A)[0.2]$, $r_4 = (B)[0.3]$, and $r_5 = (C)[0.5]$. Here, it is $\mathsf{Inc}^*(R_2) = 0.25$ with $\eta_1^* = \tau_1^* = \eta_2^* = \tau_2^* = \tau_3^* = \tau_4^* = 0$ and $\eta_3^* = 0.15$ and $\eta_4^* = 0.1$.

We show now that $\mathsf{Inc}^*$ is indeed an appropriate inconsistency measure for probabilistic knowledge bases as it satisfies the desired properties described in the previous section.

**Proposition 6.** *The function $\mathsf{Inc}^*$ satisfies (Consistency), (Inconsistency), (Monotonicity), (Super-Additivity), (Weak Independence), (Independence), (Penalty), and (Continuity).*

As mentioned before Proposition 6 also applies to the normalized inconsistency measure $\mathsf{Inc}_0^*$. This is especially true for the property (Continuity) as for $R = \emptyset$ the function $\theta_{\mathsf{Inc}_0^*, R}$ trivializes to the constant (and therefore continuous) function $\theta_{\mathsf{Inc}_0^*, R} = 0$.

## 5 Determining the Causes of Inconsistency

In Example 2 it has been shown, that the first constraint $(A \,|\, B)[1]$ had to be modified in order to restore consistency of the knowledge base $R_2$. This might tempt to make the assumption that this constraint is alone responsible for the inconsistency in $R_2$. But if one looks at $R_2$ in more depth, one can see that the situation is symmetrical in all three constraints. The optimization problem has more than one solution, as for example $\eta_2^* = 1$ also solves the inconsistency (with all other values being zero) with the same inconsistency measure of one. In fact, most inconsistent knowledge bases feature this behavior, as for the optimization problem it is best to alter as less values of the probabilistic constraints as possible. Usually, one is not only interested in determining a simple value of inconsistency, but to determine the causes of the inconsistency and ultimately to restore consistency. In Example 2 all three constraints are equally responsible for producing the inconsistency in $R_2$. In order to achieve a more in-depth analysis of the inconsistency in probabilistic knowledge bases in general, we apply the same technique as in (Hunter and Konieczny, 2006), where the *Shapley value* (Shapley, 1953) is used to determine the causes of inconsistency in propositional knowledge bases.

We go one by giving a short overview over coalition game theory and the role of the Shapley value in it, followed by the definition of the Shapley inconsistency measure on probabilistic knowledge bases.

### 5.1 Coalition Game Theory

Coalition game theory is concerned with games, where players can form coalitions in order to maximize their own payoff of the game. Let $\mathfrak{P}$ denote the power set.

**Definition 3** (Coalition Game). A *coalition game* $(N, v)$ is composed of a set of players $N \subseteq \mathbb{N}$ and a function $v : \mathfrak{P}(N) \to \mathbb{R}$ with $v(\emptyset) = 0$ and $v(S \cup T) \geq v(S) + v(T)$ for $S, T \subseteq N$ with $S \cap T = \emptyset$.

For every possible coalition $C \subseteq N$ of players in the game, the value $v(C)$ determines the payoff this coalition gets. As this payoff must be distributed on the members of $C$, every player has to evaluate for his own, which coalition to form in order to maximize his own expected payoff. Not every player has to expect the same payoff for himself, as players may be more or less important for the forming of coalitions.

**Example 4** (taken from (Hunter and Konieczny, 2006)). Let $N = \{1, 2, 3\}$ and the function $v : \mathfrak{P}(N) \to \mathbb{R}$ be defined as

$$v(\{1\}) = 1 \quad v(\{2\}) = 0 \quad v(\{3\}) = 1$$
$$v(\{1,2\}) = 10 \quad v(\{2,3\}) = 11 \quad v(\{1,3\}) = 4$$
$$v(\{1,2,3\}) = 12$$

In this game, not every player should expect the same payoff, as for instance it is more advantageous for player 1 to form a coalition with player 2 rather than with player 3 alone.

A solution to a coalition game $(N, v)$ consists of an assignment $S_i(v)$ of payoffs to each player $i \in N$, that is *fair* in the sense that every player gets as much payoff as his contribution in the grand coalition $N$ weighs. Some formal desirable properties of a solution are as follows.

(**Efficiency**) $\sum_{i \in N} S_i(v) = v(N)$

(**Symmetry**) For all $i, j \in N$, if $v(C \cup \{i\}) = v(C \cup \{j\})$ for all $C \subseteq N \setminus \{i, j\}$ then $S_i(v) = S_j(v)$

(**Dummy**) If for $i \in N$ it is $v(C) = v(C \cup \{i\})$ for all $C \subseteq N$ then it is $S_i(v) = 0$

(**Additivity**) $S_i(v + w) = S_i(v) + S_i(w)$

A solution should comprehend for the fact, that the value to be distributed among the players is the maximal value that can be achieved (Efficiency). If two players are indistinguishable by their contributions to the coalitions, they deserve the same payoff (Symmetry); if a player does not contribute to any coalition at all, his payoff should be zero (Dummy). (Additivity) describes the desired behavior of a solution if two coalition games are combined.

It can be shown (Shapley, 1953), that the *Shapley value* defined as follows is the only solution for a coalition game that satisfies (Efficiency), (Symmetry), (Dummy), and (Additivity).

**Definition 4** (Shapley Value). Let $(N, v)$ be a coalition game. The *Shapley Value* $S_i(v)$ for a player $i \in N$ is defined as

$$S_i(v) = \sum_{C \subseteq N} \frac{(|C|-1)!(|N|-|C|)!}{|N|!}(v(C) - v(C \setminus \{r\}))$$

**Example 5** ((Hunter and Konieczny, 2006)). The Shapley values for the players $1, 2, 3$ from Example 4 are

$$\begin{aligned} S_1(v) &\approx 2.83 \\ S_2(v) &\approx 5.83 \\ S_3(v) &\approx 3.3 \end{aligned}$$

### 5.2 Shapley Inconsistency Measure

We now define in accordance to (Hunter and Konieczny, 2006) a Shapley function using an inconsistency measure Inc, thus being enabled to further investigate the causes of inconsistency.

**Definition 5** (Probabilistic Shapley Inconsistency Measure). Let $K$ be a knowledge base, $r \in K$ a probabilistic constraint, and Inc an inconsistency measure. We define the *probabilistic Shapley inconsisteny measure* $S^K_{\mathsf{Inc}}(r)$ of $r$ in $K$ as

$$S^K_{\mathsf{Inc}}(r) = \sum_{C \subseteq K} \frac{(|C|-1)!(n-|C|)!}{n!}(\mathsf{Inc}(C) - \mathsf{Inc}(C \setminus \{r\}))$$

Using the probabilistic Shapley inconsistency value we can obtain more specific information about how the inconsistency is distributed on the probabilistic constraints of a knowledge base. In the following we use the inconsistency measure $\mathsf{Inc}^*$ developed in the previous section for the application of the Shapley inconsistency measure.

**Example 6.** Consider again the knowledge base $R_1 = \{r_1, r_2, r_3, r_4\}$ with $r_1 = (A \,|\, \overline{B})[0.8]$, $r_2 = (A \,|\, B)[0.6]$, $r_3 = (B)[0.5]$, and $r_4 = (A)[0.2]$ from Example 1 with $\mathsf{Inc}^*(R_1) = 0.5$. There it is $S^{R_1}_{\mathsf{Inc}^*}(r_1) \approx 0.15$, $S^{R_1}_{\mathsf{Inc}^*}(r_2) \approx 0.117$, $S^{R_1}_{\mathsf{Inc}^*}(r_3) \approx 0.05$, and $S^{R_1}_{\mathsf{Inc}^*}(r_4) \approx 0.183$. The distribution of the Shapley values indicates that the constraint $r_4 = (A)[0.2]$ is mostly responsible for the inconsistency in $R_1$ and $r_3 = (B)[0.5]$ is less responsible. This can be justified as both rules $r_1$ and $r_2$ describe an influence of $B$ on $A$ and – assuming that the knowledge base describes causal rather that diagnostic information – thus state that $B$ is more entrenched or more basic than $A$. Thus, rule $r_4$ that gives a probability of $A$ not conditioned on anything else, is most dangerous for consistency.

**Example 7.** Consider again the knowledge base $R_2 = \{r_1, r_2, r_3\}$ with $r_1 = (A \,|\, B)[1]$, $r_2 = (B)[1]$, and $r_3 = (A)[0]$ from Example 2 with $\mathsf{Inc}^*(R_2) = 1$. There it is $S^{R_2}_{\mathsf{Inc}^*}(r_1) \approx 0.33$, $S^{R_2}_{\mathsf{Inc}^*}(r_2) \approx 0.33$, and $S^{R_2}_{\mathsf{Inc}^*}(r_3) \approx 0.33$. Here it is clear, that all three probabilistic constraints are equally responsible for the inconsistency in $R_2$. This reflects the intuition described above.

**Example 8.** Consider again the knowledge base $R_3 = \{r_1, r_2, r_3, r_4, r_5\}$ with $r_1 = (A \,|\, C)[0.7]$, $r_2 = (B \,|\, \overline{C})[0.8]$, $r_3 = (A)[0.2]$, $r_4 = (B)[0.3]$, and $r_5 = (C)[0.5]$ from Example 3 with $\mathsf{Inc}^*(R_3) = 0.25$. There it is $S^{R_3}_{\mathsf{Inc}^*}(r_1) \approx 0.062$, $S^{R_3}_{\mathsf{Inc}^*}(r_2) \approx 0.045$, $S^{R_3}_{\mathsf{Inc}^*}(r_3) \approx 0.062$, $S^{R_3}_{\mathsf{Inc}^*}(r_4) \approx 0.045$, and $S^{R_3}_{\mathsf{Inc}^*}(r_5) \approx 0.036$.

The probabilistic Shapley inconsistency measure satisfies the same properties as the Shapley value due to its direct application on an inconsistency measure.

**Proposition 7.** *If $\mathsf{Inc}$ is an inconsistency measure that satisfies (Consistency) and (Super-Additivity), then the probabilistic Shapley inconsistency measure $S_{\mathsf{Inc}}$ satisfies*

**(Efficiency)** $\sum_{r \in R} S^R_{\mathsf{Inc}}(r) = \mathsf{Inc}(R)$

**(Symmetry)** *For all $r, r' \in R$, if $\mathsf{Inc}(R' \cup \{r\}) = \mathsf{Inc}(R' \cup \{r'\})$ for all $R' \subseteq R \setminus \{r, r'\}$ then $S^R_{\mathsf{Inc}}(r) = S_{\mathsf{Inc}}(r')$*

**(Dummy)** *If for $r \in R$ it is $\mathsf{Inc}(R') = \mathsf{Inc}(R' \cup \{r\})$ for all $R' \subseteq R$ then it is $S^R_{\mathsf{Inc}}(r) = 0$*

The property (Additivity) was neglected as it does not make sense for our application, cf. (Hunter and Konieczny, 2006). From Proposition 7 it follows directly

**Corollary 1.** *The probabilistic Shapley inconsistency measure $S_{\mathsf{Inc}^*}$ satisfies (Efficiency), (Symmetry), and (Dummy).*

Using the inconsistency measure $\mathsf{Inc}^*$ and the Shapley inconsistency measure $S^R_{\mathsf{Inc}^*}$ the knowledge engineer can support his efforts to restore consistency in a probabilistic knowledge base $R$. The solutions of the optimization problem $OPT_R$ describe the minimal adjustments to be made in order to restore consistency. By considering the Shapley inconsistency measures $S^R_{\mathsf{Inc}^*}$ for the constraints in $R$ one can select the most appropriate solution and modify the knowledge base accordingly. We will formalize this approach of restoring consistency using inconsistency measures in an upcoming paper.

## 6 Summary and Discussion

We developed an inconsistency measure on conditional probabilistic knowledge bases and showed that this measure satisfies several desirable properties. We went on by using this measure and the well-known Shapley value to define a more sophisticated measure that gives the knowledge engineer the means to restore consistency of a knowledge base by identifying the causes for the inconsistency. Due to its heritage from the Shapley value the Shapley inconsistency measure satisfies several nice properties.

The examples in this paper were computed by a prototypical implementation of the function $\mathsf{Inc}^*$ and $S^R_{\mathsf{Inc}^*}$ in Java which uses the free optimization software OpenOpt[2] to solve the optimization problems of type $OPT_R$.

The inconsistency measure described here and inconsistency measures in general are useful when dealing with data from heterogenous sources as in information fusion or belief revision (Bloch and Hunter, 2001; Alchourrón et al., 1985; Kern-Isberner and Rödder, 2003). They can point at discrepancies in the data representation and give hints on possible fixes within the model. Many expert systems in highly-crucial application areas like medical diagnosis or fraud detection demand to draw consistent conclusions but need to use distorted and ambiguous information at the same time. Actually, we are currently investigating the possibilities of applying information engineering techniques as the described inconsistency measure and information fusion in the field of fraud detection in the annual audit relating to commercial law. For future work we also plan to extend and apply the work reported here on relational probabilistic knowledge bases.

---

[2] http://openopt.org/

**Acknowledgments.** The research reported here was partially supported by the Deutsche Forschungsgemeinschaft under grant KE 1413/2-1.